\documentclass[10pt,twocolumn,letterpaper]{article}

\usepackage[pagenumbers]{wacv} 

\usepackage{graphicx}
\usepackage{amsmath}
\usepackage{amssymb}
\usepackage{booktabs}
\usepackage{amsmath,graphicx,multirow,multicol,adjustbox,lipsum}
\usepackage{tabularx}
\usepackage[table]{xcolor}
\usepackage{caption}

\usepackage[pagebackref,breaklinks,colorlinks]{hyperref}
\newcommand{\tabref}[1]{Tab.~\ref{#1}}

\newcommand{\figref}[1]{Fig.~\ref{#1}}

\usepackage[capitalize]{cleveref}
\crefname{section}{Sec.}{Secs.}
\Crefname{section}{Section}{Sections}
\Crefname{table}{Table}{Tables}
\crefname{table}{Tab.}{Tabs.}

\def\wacvPaperID{2136} 
\def\confName{WACV}
\def\confYear{2025}

\title{EfficientHuman: Efficient Training and Reconstruction of Moving Human using Articulated 2D Gaussian}
\author{Hao Tian$^{1,3}$, Rui Liu$^{2}$, Wen Shen$^{1,3}$, Yilong Hu$^{1,3}$, Zhihao Zheng$^{1,3}$ and Xiaolin Qin$^{1,3,*}$
\\
$^1$Chengdu Institute of Computer Applications, Chinese Academy of Sciences\\$^2$Minzu University of China \\$^3$University of Chinese Academy of Sciences \\
}

\begin{document}

\twocolumn[{%
        \renewcommand\twocolumn[1][]{#1}%
        \maketitle
        \begin{center}
	   \centering
	   \includegraphics[width=0.95\textwidth]{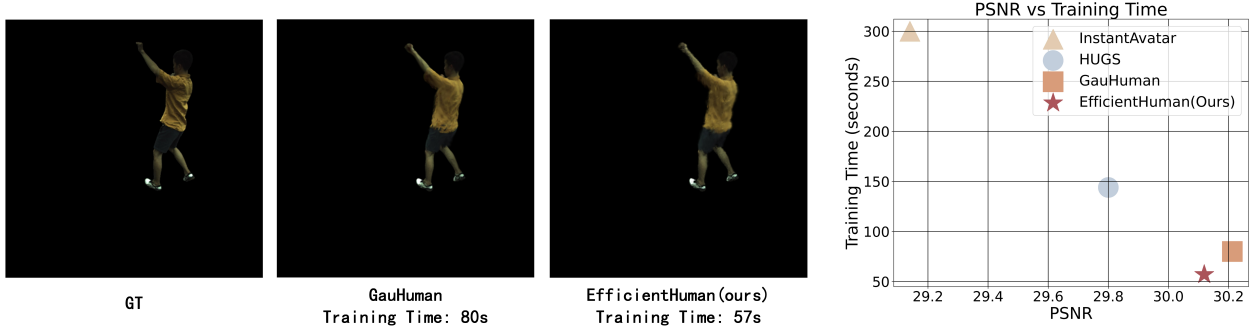}
	   \vspace*{-2mm}
	   \captionof{figure}{EfficientHuman receives the monocular videos to model a 3D human with \textbf{currently the shortest training time} and ensuring \textbf{high-quality rendering}. The left images provide a comparative analysis of the training efficiency and image quality between the current state-of-the-art and our proposed EfficientHuman method, our work not only guarantees the integrity of clothing edges but also achieves this with a significantly reduced training time. The right comparison plot demonstrates the Peak Signal-to-Noise Ratio (PSNR) and Training Time, highlighting the superior performance of EfficientHuman in ensuring PSNR within a significantly reduced training duration.}
	\label{fig:fig1}
	\vspace{0.5\baselineskip}
    \end{center}%
}]
%
\renewcommand{\thefootnote}{}
\begin{abstract}
3D Gaussian Splatting (3DGS) has been recognized as a pioneering technique in scene reconstruction and novel view synthesis. Recent work on reconstructing the 3D human body using 3DGS attempts to leverage prior information on human pose to enhance rendering quality and improve training speed. However, it struggles to effectively fit dynamic surface planes due to multi-view inconsistency and redundant Gaussians. This inconsistency arises because Gaussian ellipsoids cannot accurately represent the surfaces of dynamic objects, which hinders the rapid reconstruction of the dynamic human body. Meanwhile, the prevalence of redundant Gaussians means that the training time of these works is still not ideal for quickly fitting a dynamic human body.
To address these, we propose EfficientHuman, a model that quickly accomplishes the dynamic reconstruction of the human body using Articulated 2D Gaussian while ensuring high rendering quality. The key innovation involves encoding Gaussian splats as Articulated 2D Gaussian surfels in canonical space and then transforming them to pose space via Linear Blend Skinning (LBS) to achieve efficient pose transformations. Unlike 3D Gaussians, Articulated 2D Gaussian surfels can quickly conform to the dynamic human body while ensuring view-consistent geometries. Additionally, we introduce a pose calibration module and an LBS optimization module to achieve precise fitting of dynamic human poses, enhancing the model's performance. Extensive experiments on the ZJU-MoCap dataset demonstrate that EfficientHuman achieves rapid 3D dynamic human reconstruction in less than a minute on average, which is 20 seconds faster than the current state-of-the-art method, while also reducing the number of redundant Gaussians. 

\end{abstract}
\section{Introduction}
\label{sec:intro}
Efficiently generating detailed 3D models of human performers tailored to user requirements is crucial for applications in augmented reality (AR), virtual reality (VR), video games, and film production. Methods that learn 3D human images from sparse video graphs or individual images using implicit NeRF representations \cite{R5} have been shown to achieve high-quality 3D human modeling. However, training these methods typically requires significant time and computational resources. For example, NeRF-based human modeling methods \cite{R1, R20, R3, R4} often require several hours of fine-tuning in advance to achieve satisfactory 3D human representations. To alleviate this, recent research \cite{R6, R7, R8} has focused on developing efficient methods for the three-dimensional characterization of the human body, aiming to reduce the required training time. Due to the inefficiency of backward mapping-based ray-rendering algorithms \cite{R5}, the training time remains substantial.

Given the superior speed of 3D Gaussian Splatting (3DGS) \cite{R9} in reconstructing static scenes compared to NeRF \cite{R5}, recent work \cite{R12, R13} has proposed using 3DGS for dynamic human body reconstruction. By adding human prior information like SMPL \cite{R10}, these methods require only a limited set of 3D Gaussians to achieve the same effect as traditional NeRF-based human models \cite{R1, R20, R3, R4}, the bottleneck in quickly fitting the dynamic human surface with Gaussian ellipsoids results in a large number of redundant Gaussians. This redundancy indicates that there is still considerable room for improving the convergence speed.

To prune the number of redundant Gaussians, the 2D Gaussian \cite{R14} can directly model static surface planes, as opposed to the 3D Gaussian \cite{R9}, which covers the surface. When it is taken to reconstruct a dynamic human body, the promising approach will face two challenges: 1) \textbf{2D Gaussian is designed to focus on static surface modeling}, and no existing method integrates 2D Gaussians with human motion information to achieve the rapid reconstruction of a clear dynamic human body. 2) \textbf{Inaccuracy in the human body's pose will also inevitably lead to 2D Gaussian misalignment}, resulting in 2D Gaussian can't ensure high-quality rendering. Therefore, it is crucial to develop a way to constrain Gaussians within the Gaussian Splatting framework to accelerate the training process and minimize the number of redundant Gaussians, while ensuring high-quality rendering. 

To handle the two challenges, we propose EfficientHuman, which shows considerable promise for reconstructing a dynamic human body through \textbf{Articulated 2D Gaussian}. In this approach, monocular videos serve as the input. After initializing the Gaussians, the Gaussian splats and SMPL human priors are parametrically encoded as Articulated 2D Gaussians in canonical space. Finally, these Articulated 2D Gaussians are transformed into pose space using Linear Blend Skinning (LBS) \cite{R10}. It is important to note that directly using human prior information during LBS is not advisable because the SMPL prior is not specially designed for the representation of Gaussian, which still leads to inaccuracy of LBS transformation due to compatibility. To address this, we design a \textbf{pose calibration} module to refine the T-pose of the canonical human body and an \textbf{LBS optimization} module to fine-tune the LBS weights for more accurate transformations. Through the series of control methods mentioned above, our proposed EfficientHuman model can efficiently manage Articulated 2D Gaussians to achieve dynamic human reconstruction. To our knowledge, this is the first model to apply 2D Gaussians for 3D human body reconstruction. We achieve an average training time of \textbf{less than 60 seconds}, which is \textbf{20 seconds faster than the current state-of-the-art} \cite{R12}, using only \textbf{1.2K iterations}. Additionally, we \textbf{prune redundant Gaussians} while ensuring \textbf{high-quality rendering}. Our contributions include:

1.\hspace{2pt}We propose a novel 3D human body reconstruction model, EfficientHuman, which employs \textbf{Articulated 2D Gaussians} to efficiently model dynamic humans and represent the human body accurately in 3D space.

2.\hspace{2pt}We design a \textbf{pose calibration} module and an \textbf{LBS optimization} module to enable more precise transformations to the target pose space. Additionally, we introduce a regularization loss to enhance performance efficiently.

3.\hspace{2pt}Experiments on the ZJU-MoCap dataset \cite{R37} demonstrate that EfficientHuman can perform rapid 3D dynamic human reconstruction in less than a minute on average, achieving a training speed that is $\approx \textbf{30\%}$ faster than the current state-of-the-art, while reducing redundant Gaussians and maintaining high-quality rendering.

\section{Related work}
\label{sec:format}
\textbf{3D Human Reconstruction.}
Rapid reconstruction of 3D dynamic humans from monocular videos remains a significant challenge. Traditional methods rely on multi-view stereo \cite{R15, R16} or depth fusion \cite{R17, R18, R19} to restore the geometry of a moving human, but this approach requires numerous cameras and sensors to capture images and depth information from various perspectives. To alleviate these requirements, some approaches \cite{R20, R21, R22, R4} incorporate prior human information. The prior information is often collected from different pre-trained datasets, which lack diversity, making it difficult to accurately reconstruct a freely moving human body with complex motions.
Another approach \cite{R24, R25, R26, R37, R28, R29} involves using implicit neural fields to represent the features of a dynamic human body in 3D space from sparse-view video or a single image. However, the extensive training time required by this method limits its practical application in real-world scenarios.

\textbf{Boosting Neural Rendering Performance.}
With the advent of Neural Radiance Fields (NeRF) \cite{R5}, researchers have recognized the significant time consumption required for training. As a result, various methods have been developed to accelerate the NeRF's speed. During rendering, techniques such as voxel grids \cite{R8, R43, R44, R45, R47, R48, R49} and point-based representations \cite{R50, R51, R52, R53, R54} are employed to reduce the number of queries to the NeRF multilayer perceptron, thereby optimizing NeRF's speed.
In the field of human reconstruction, some methods utilize UV maps \cite{R55, R56, R57} or voxel grids \cite{R58}, but these techniques are primarily effective for reconstructing static scenes. When applied to dynamic scenes, notably the restoration of a dynamic human body, these methods entails substantial training time requirements.

\textbf{3D Gaussian Splatting.}
3D Gaussian Splatting (3DGS) \cite{R9} has received widespread attention since its proposal, this method demonstrates high-speed training and high-fidelity rendering. Although some researchers \cite{R12, R13} utilize 3DGS to model a 3D human with competitive training speed relative to NeRF-based human models \cite{R1, R20, R3, R4}. However, 3D Gaussian can only cover the surface of a dynamic human body, the inherent feature will generate a large number of Gaussians to fit the dynamic human body, which suggests that there is still significant room for improvement in training speed.

Our work efficiently reconstructs a human body in motion, minimizing the final Gaussian count. We utilize the Articulated 2D Gaussian to manage the 2D Gaussian \cite{R14}, using the SMPL model \cite{R10} as prior knowledge for learning a deformation space crucial for animation control. To achieve precise reproduction of human motion, we incorporate a pose calibration module and a Linear Blend Skinning \cite{R10} optimization module.

\section{Method}
\label{sec:method}
Our work begins with monocular video as the input. After initializing the Gaussians, the Gaussian splats are parametrically encoded as Articulated 2D Gaussians in canonical space. And then these Articulated 2D Gaussians are decomposed into encoded 3D positions \(\boldsymbol{p}_c^e\) in canonical space and initial weights \(\boldsymbol{\theta}_T\) of the T-pose. The \(\boldsymbol{p}_c^e\) is fine-tuned through the Linear Blend Skinning (LBS) \cite{R10} optimization module, while \(\boldsymbol{\theta}_T\) is corrected by the pose calibration module. Finally, we update the weight matrix \(\boldsymbol{W}_k\), rotation \(\boldsymbol{R}\) and translation \(\boldsymbol{T}\), transforming the Articulated 2D Gaussian into pose space using LBS.

\begin{figure*}[!]
    \centering
    \includegraphics[width=\textwidth]{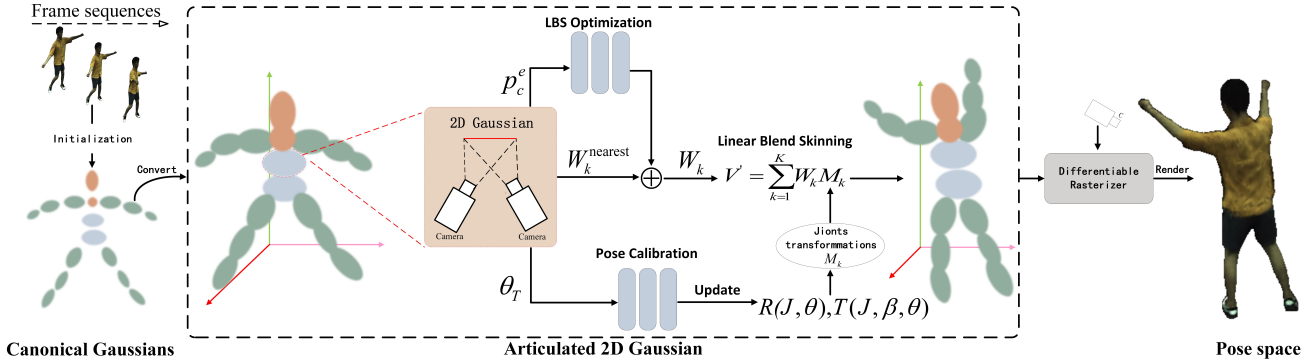}
    \caption{\textbf{EfficientHuman method overview.} Our pipeline is initiated with the ingestion of sequential video frames, followed by the initialization of Gaussians within a predefined canonical space. These Gaussians are subsequently encoded and articulated into a coherent set of 2D Gaussian surfels with human prior, embodying the subject in an anatomical T-pose. Notably, the 2D Gaussian surfels in our method feature negligible gaps, in contrast to the more substantial separations found between Gaussian ellipsoids. The transformation and control of these surfels are directed by the optimized parameters \(\theta_T\) and \(p_c^e\), which are optimized through modules of pose calibration and Linear Blend Skinning (LBS) optimization, respectively. Finally, the 2D Gaussian surfels are mapped into the pose space of the target camera view using LBS.
   }
    \label{fig:fig2}
\end{figure*}
\subsection{Preliminary}
\label{ssec:subhead}

\textbf{3D Gaussian Splatting.}
3D Gaussian Splatting (3DGS)\cite{R9} represents 3D scene features using five parameters: position, transparency, spherical harmonic coefficients, rotation, and scaling. The 3D Gaussian is first initialized using Structure-from-Motion (SFM) \cite{R36, R39} or random initialization, while a differentiable parallelizable rasterization technique is applied to render the image. Specifically, 3DGS explicitly defines each 3D Gaussian ellipsoid in space using a covariance matrix \(\boldsymbol{\Sigma}\) and the position vector \(\boldsymbol{\rho}\), as shown in the following equation:
\begin{equation}
\mathcal{G}(\boldsymbol{x})=\frac{1}{(2 \pi)^{\frac{3}{2}}|\boldsymbol{\Sigma}|^{\frac{1}{2}}} \mathrm{exp}{(-\frac{1}{2}(\boldsymbol{x}-\boldsymbol{\rho})^{\top} \boldsymbol{\Sigma}^{-1}(\boldsymbol{x}-\boldsymbol{\rho}))}
\end{equation}

Here the \(\boldsymbol{\Sigma}\) matrix can be decomposed into a rotation matrix $\boldsymbol{R}$ and a scaling matrix $\boldsymbol{S}$ by \(\boldsymbol{\Sigma} = \boldsymbol{RSS}^{\top} \boldsymbol{R}^{\top}\). The 3D Gaussian is then transformed from world coordinates to camera coordinates by the word-to-camera transformation matrix $\boldsymbol{w}$ and the projection matrix $\boldsymbol{\mathcal{J}}$.
\begin{equation}
\boldsymbol{\Sigma}^{\prime}=\boldsymbol{\mathcal{J} w \Sigma w}^{\top}\boldsymbol{\mathcal{J}}^{\top}
\end{equation}

The 3D Gaussian is projected into 2D space and the color \(c_i\) and density \(\alpha_i\) distribution of overlapping Gaussians on each pixel is counted, and the final color of the pixel is obtained by sorting the Gaussians on each pixel using blending $N$. The formula is expressed as follows:

\begin{equation}
\hat{C}=\sum_{i \in N} c_i \alpha_i \prod_{j=1}^{i-1}\left(1-\alpha_i\right)
\end{equation}

While this method enables end-to-end training and rendering via differentiable rasterization, relying solely on 3DGS for reconstructing a dynamic human body can lead to slow convergence and Gaussian bias. This is due to the overproduction of Gaussian ellipsoids required to accurately capture the free movement of a human body in space.

\textbf{2D Gaussian Splatting.}
The introduction of 2D Gaussian Splatting (2DGS) \cite{R14} facilitates 3D modeling by flattening a 3D Gaussian ellipsoid \cite{R9} into a 2D Gaussian surfel, this transformation simplifies the representation by embedding a 2D Gaussian within a 3D space. In essence, the 3D Gaussian is converted from a three-dimensional coordinate system to a surfel defined by a two-dimensional coordinate system, with the normal line direction determined by the steepest density gradient. This design enhances the model's ability to accurately align with the object's surface.

The alignment of the 2D Gaussian on the surface of the object is mainly done by the center point of the Gaussian \(\boldsymbol{p}_k\), two orthogonal tangent vectors \(\boldsymbol{t}_u\) and \(\boldsymbol{t}_v\), and a scaling matrix \(\boldsymbol{S}\) consisting of the scaling vectors \(\boldsymbol{s}_u\) and \(\boldsymbol{s}_v\) corresponding to the two tangent vectors, where \(\boldsymbol{S}\) is used to control the 2D Gaussian size change. The original normal is declared as \(\boldsymbol{t}_w = \boldsymbol{t}_u \times \boldsymbol{t}_v\), the rotation matrix as \(\boldsymbol{R}_{3\times3} = [\boldsymbol{t}_u, \boldsymbol{t}_v, \boldsymbol{t}_w]\), and the scaling matrix as \(\boldsymbol{S}_{3\times3} = [\boldsymbol{s}_u, \boldsymbol{s}_v, 0]\). The tangent plane parameterization in real space can be represented as:
\begin{equation}
\begin{aligned}
\boldsymbol{P}(u, v) & = \boldsymbol{p}_k+\boldsymbol{s}_u \boldsymbol{t}_u u+\boldsymbol{s}_v \boldsymbol{t}_v v = \mathbf{H}(u, v, 1,1)^{\top} \\
\end{aligned}
\end{equation}
\begin{equation}
\begin{aligned}
\text { where } \mathbf{H} & =\left[\begin{array}{cccc}
\boldsymbol{s}_u\boldsymbol{t}_u & \boldsymbol{s}_v \boldsymbol{t}_v & 0 & \boldsymbol{p}_k \\
0 & 0 & 0 & 1
\end{array}\right]=\left[\begin{array}{cc}
\boldsymbol{RS} & \boldsymbol{p}_k \\
0 & 1
\end{array}\right]
\end{aligned}
\end{equation}

Here $\mathbf{H}$ is a \(4\times4\) matrix that acts as a homogeneous transformation of the 2D Gaussian, and if \(\boldsymbol{u}=(u,v)\) is used as a point in \(uv\) space, then each 2D Gaussian is represented as a similar form of 3D Gaussian as:
\begin{equation}
\mathcal{G}(\boldsymbol{u})=\exp \left(-\frac{u^2+v^2}{2}\right)
\end{equation}

Concurrently, each 2D Gaussian surfel is also characterized by a color $c$, derived from transparency \(\alpha_i\) and spherical harmonic functions, as well as a covariance matrix given by \(\boldsymbol{\Sigma} = \boldsymbol{RSS}^{\top}\boldsymbol{R}^{\top}\). These parameters negate the necessity for substantial modifications to the 2DGS optimization process, and the density adjustment method can adopt the optimization approach utilized in 3D Gaussian Splatting \cite{R9}.

\subsection{Articulated 2D Gaussian}
\label{ssec:subhead} 
In the field of 3D human reconstruction, researchers \cite{R12, R13} have utilized 3D Gaussian Splatting (3DGS) \cite{R9} to expedite training and enhance fidelity. However, 3DGS-based methods encounter a significant challenge: the redundancy of Gaussians not only fails to enhance reconstruction quality but also prolongs the training duration. Moreover, incorporating multilayer perceptrons for pose recognition further extends the duration of the training. Thus, the optimization process necessitates targeted and strategic enhancements to accelerate training. While 2D Gaussian Splatting \cite{R14} is adept at static surface modeling, it falls short in capturing the dynamics of human surfaces. To address this, we introduce the Articulated 2D Gaussian, which leverages prior knowledge of human anatomy to guide the 2D Gaussian fitting. By conforming directly to the dynamic human surface using 2D Gaussian surfels, we achieve a reduction in both training time and the number of Gaussians. We also propose two optimization modules to ensure high rendering quality and to accurately calibrate the human body from a canonical to a pose space. Additionally, our model's performance is bolstered by the inclusion of the \(\mathcal{L}_{mask}\) loss, with a minimal impact on training time.

As shown in \figref{fig:fig2}, we initialize a series of Gaussians by processing the input frame stream from the monocular video, and then parameterize these Gaussians in canonical space as Articulated 2D Gaussian surfels using SMPL \cite{R10} information. The prior information from the SMPL is parameterized by $V = (\boldsymbol{\beta},\boldsymbol{\theta}, \boldsymbol{J}, \boldsymbol{W})$, where $V$ denotes the positions of the vertex, $\boldsymbol{J}$ represents a sequence of joint transformations, and $\boldsymbol{W}$ is the skinning weight matrix. With these parameters established, the body's shape and pose are adjustable via $\boldsymbol{\beta}$ and $\boldsymbol{\theta}$. Each Articulated 2D Gaussian in the canonical space has the following parameters: \(\boldsymbol{\theta}_T\) is the initial weights of vertices in the T-pose, \(\boldsymbol{W}_k^{\mathrm{nearest}}\) is the nearest weight associated with the \(k\)-th joint, \(\boldsymbol{p}_c\) and \(\boldsymbol{\Sigma}_c\) represent the initial 3D position without position encoding and covariance matrix of T-pose in the canonical space, respectively. An initial T-pose of the Articulated 2D Gaussian body is constructed using these parameters integrated with the SMPL prior. Finally, these surfels are mapped into the pose space using Linear Blend Skinning (LBS) \cite{R10}. 
\setlength{\baselineskip}{0.95\baselineskip}
\begin{equation}
\boldsymbol{p}_c^e=(\small\sin (2^0 \pi \boldsymbol{p}_c), \small\cos (2^0 \pi \boldsymbol{p}_c), \ldots, \small\sin (2^L \pi \boldsymbol{p}_c), \small\cos (2^L \pi \boldsymbol{p}_c))
\end{equation}

\begin{equation}
\boldsymbol{p}_k=\boldsymbol{R}(\boldsymbol{J}, \boldsymbol{\theta}) \boldsymbol{p}_c^e+\boldsymbol{T}(\boldsymbol{J}, \boldsymbol{\beta}, \boldsymbol{\theta})
\end{equation}
\begin{equation}
\boldsymbol{\Sigma}_k = \boldsymbol{R}(\boldsymbol{J},\boldsymbol{\theta})\boldsymbol{\Sigma}_c\boldsymbol{R}(\boldsymbol{J},\boldsymbol{\theta})^{\top}
\end{equation}

where \(\boldsymbol{p}_k\),\(\boldsymbol{\Sigma}_k\) are the final position of the 2D Gaussian and covariance matrix in pose space. The final 3D position of the canonical space \(\boldsymbol{p}_c^e\) is generated by position encoding \cite{R30, R31, R32, R33, R34, R35}, this encoding can capture spatial position information at different frequencies, which determines the vector dimension after encoding. $L$ is the number of layers of position encoding, $k$ stands for the joint position in $\boldsymbol{J}$.

\begin{equation}
\boldsymbol{R}(\boldsymbol{J}, \boldsymbol{\theta})=\sum_{k=1}^K \boldsymbol{W}_k \boldsymbol{R}_k(\boldsymbol{J}, \boldsymbol{\theta})
\end{equation}
\begin{equation}
\boldsymbol{T}(\boldsymbol{J}, \boldsymbol{\beta}, \boldsymbol{\theta})=\sum_{k=1}^K \boldsymbol{W}_k \boldsymbol{T}_k(\boldsymbol{J}, \boldsymbol{\beta}, \boldsymbol{\theta})
\end{equation}

\(\boldsymbol{R}(\boldsymbol{J},\boldsymbol{\theta})\) and \(\boldsymbol{T}(\boldsymbol{J},\boldsymbol{\beta},\boldsymbol{\theta})\) are the rotation matrix and translation vector, respectively. \(\boldsymbol{R}_k(\boldsymbol{J},\boldsymbol{\theta})\) and \(\boldsymbol{T}_k(\boldsymbol{J},\boldsymbol{\beta},\boldsymbol{\theta})\)  represent the transformation matrix and translation vector of the \(k\)-th joint, respectively. \(\boldsymbol{W}_k\) is the weight matrix after optimization by the LBS optimization module, and $K$ is the upper limit of the number of $k$.

Specifically, we use \(\boldsymbol{R}(\boldsymbol{J},\boldsymbol{\theta})\) and \(\boldsymbol{T}(\boldsymbol{J},\boldsymbol{\beta},\boldsymbol{\theta})\) to guide the rotation and translation of the Articulated 2D Gaussian surfel. Meanwhile, \(\boldsymbol{R}(\boldsymbol{J},\boldsymbol{\theta})\) and \(\boldsymbol{T}(\boldsymbol{J},\boldsymbol{\beta},\boldsymbol{\theta})\) is applied by LBS, as seen in the following equation:
\begin{equation}
\begin{aligned}
\boldsymbol{V}^{\prime}=\sum_{k=1}^K \boldsymbol{W}_k \mathbf{M}_k
\end{aligned}
\end{equation}
\begin{equation}
\begin{aligned}
\text { where } \mathbf{M}_k=\left[\begin{array}{cc}
\boldsymbol{R}_k(\boldsymbol{J},\boldsymbol{\theta}) & \boldsymbol{T}_k(\boldsymbol{J},\boldsymbol{\beta},\boldsymbol{\theta}) \\
0 & 1
\end{array}\right]
\end{aligned}
\end{equation}

Here, $\mathbf{M}_k$ represents the transformation matrix of the \(k\)-th joint, which includes rotation, translation and scaling. $\boldsymbol{V}^{\prime}$ is the new position of the vertex after transformation, and $K$ is the upper limit of the joints.
Afterward, formula (6) is used to generate \(\mathcal{G}(\boldsymbol{u})\), and alpha blending is applied to integrate the alpha-weighted appearance from front to back.
\begin{equation}
c(\boldsymbol{u})=\sum_{i=1}^I c_i \alpha_i \mathcal{G}_i(\boldsymbol{u}) \prod_{j=1}^{i-1}\left(1-\alpha_j \mathcal{G}_j(\boldsymbol{u})\right)
\end{equation}

where \(i\) is the index of each articulated 2D Gaussian surfel, \(\alpha_i\) is value of alpha, \(c_i\) is the appearance of view dependent. By passing light through different surfels, the final output is the appearance of the current pixel.

\subsection{Linear Blend Skinning (LBS) Optimization}
\label{ssec:subhead} 
Statistical model of human SMPL \cite{R10} may not capture all the nuances of individual body shapes, leading to deviations from the actual shape. This can cause Gaussians to be placed inaccurately after Linear Blend Skinning (LBS) \cite{R10}, especially in regions where fine details and hand motions are crucial. Ensuring an accurate translation of SMPL's parametric data into the Gaussian Splatting framework \cite{R9, R14} requires special optimization.

To ensure that Articulated 2D Gaussians correctly represent the human body in various poses, the LBS optimization module is applied before the LBS transformation. This module aims to optimize the LBS transformation matrix to guide the rotation and translation of the 2D Gaussians by constraining the LBS misalignment. In other words, the Gaussians are made to move and rotate in accordance with the underlying skeletal structure, akin to how mesh vertices deform in response to skeletal transformations.
\begin{equation}
    \small \boldsymbol{W}_k = \frac{\mathrm{exp}(\log(\boldsymbol{W}_k^{\mathrm{nearest}}+10^{-8}) + \mathrm{opt}_{\mathrm{LBS}}(\boldsymbol{p}_c^e)[k])}{\sum_{k=1}^K\mathrm{exp}(\log(\boldsymbol{W}_k^{\mathrm{nearest}}+10^{-8})+\mathrm{opt}_{\mathrm{LBS}}(\boldsymbol{p}_c^e)[k])}
\end{equation}

Specifically, \(\mathrm{opt}_{\mathrm{LBS}}(\boldsymbol{\cdot})\) is a multilayer perceptual fine-tuned to \(\boldsymbol{p}_c^e\). The parameter \(\boldsymbol{W}_k^{\mathrm{nearest}}\) is set to be the nearest LBS weight of the $k$-th joint in the SMPL model. To prevent zero or negative values, a small positive increment of \(10^{-8}\) is added. Additionally, logarithmic transformation is applied to stabilize the weight ratios and avoid numerical instability. The outcomes from \(\mathrm{opt}_{\mathrm{LBS}}(\boldsymbol{\cdot})\) are aggregated and subsequently normalized to ensure that the sum of all \(\boldsymbol{W}_k\) weights equals 1.

These operations correct the biased LBS weight matrix to constrain Gaussian misalignment, addressing the inevitable pose transformation inaccuracies introduced by SMPL as prior information.

\subsection{Pose Calibration}
\label{ssec:subhead} 
Since Gaussian Splatting \cite{R9, R14} relies on accurate 3D positions and orientations derived from the SMPL model \cite{R10}, any errors in the SMPL output will propagate through the pipeline, leading to noticeable misalignment or distortion in the final 2D Gaussian representation.
The pose calibration module is used to standardize the T-pose of a 3D human body, which is conducive to a better realization of accurate Linear Blend Skinning(LBS) \cite{R10}.
\begin{equation}
\boldsymbol{\theta} = \boldsymbol{\theta}_T + \mathrm{\Delta}\boldsymbol{\theta}
\end{equation}
\begin{equation}
\mathrm{{\Delta}}\boldsymbol{\theta} = \mathrm{MLP}_{\mathrm{pose}}(\boldsymbol{\theta}_T)
\end{equation}

\(\boldsymbol{\theta}_T\) is the initial weights of the T-pose, and then \(\mathrm{\Delta}\boldsymbol{\theta}\) obtained by fine-tuning \(\boldsymbol{\theta}_T\) via \(\mathrm{MLP}_{\mathrm{pose}}(\boldsymbol{\cdot})\) is added to get the final pose parameter \(\boldsymbol{\theta}\). 
\begin{figure*}[!]
    \centering
    \includegraphics[width=\textwidth]{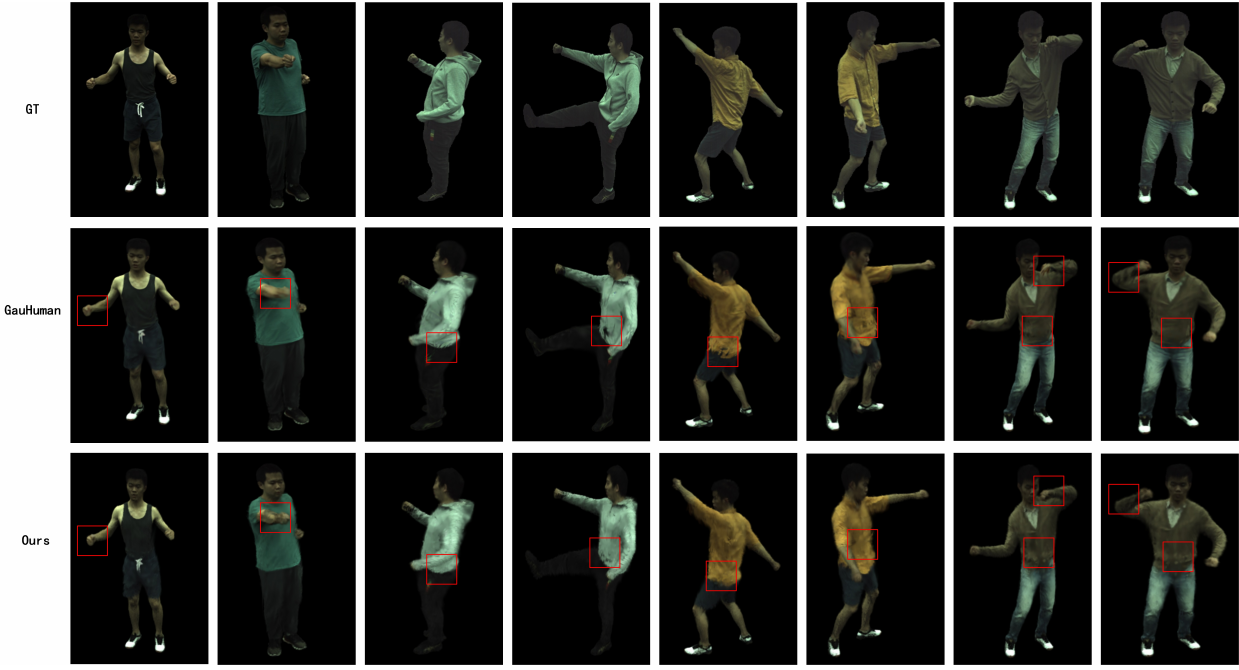}
    \caption{Qualitative comparison of our approach and the baseline method on ZJU-MoCap, highlighting six distinct actions. The rendering quality of our method is comparable with the baseline method, and Figure 3 also illustrates the improvement of our reconstruction method over the baseline, particularly in the red-boxed area of row three and columns one to eight. This is evident in the preservation of fine details, such as the clothing button and edges, as well as the improved recovery and rendering of human arm contours. Zoom in for the best view.
}
    \label{fig:fig3}
\end{figure*}

By fine-tuning the initial T-pose weights \(\boldsymbol{\theta}_T\) through the pose calibration process, denoted as \(\mathrm{MLP}_{\mathrm{pose}}(\boldsymbol{\theta}_T)\), the module adjusts the pose parameters \(\boldsymbol{\theta}\) to reflect the specific posture more accurately. This fine-tuning minimizes discrepancies between the expected and actual positions of the 2D Gaussians, thereby reducing potential misalignment or distortion. Consequently, the final Articulated 2D Gaussian representation better aligns with the intended human pose, leading to improved accuracy and visual consistency.

Articulated 2D Gaussians enable efficient dynamic surface modeling compared to 3D Gaussians, with editable parameters encoding pose information into 2D Gaussian surfel properties \cite{R14}. Our proposed EfficientHuman method utilizes Articulated 2D Gaussians to render a 3D human with quality approximating state-of-the-art results.

\subsection{Optimization and Detail}
\label{ssec:subhead} 
3D Gaussian Splatting \cite{R9} can be initialized using Structure-from-Motion (SFM) \cite{R36, R39} or random initialization. However, such initializations may lose the spatial distribution of the human body and are not well-suited for high degrees of freedom in human motion. The SMPL model \cite{R10}, a statistical model based on extensive human body data, provides valuable prior information. The 6890 vertices from SMPL is utilized to initialize the Gaussians, which represent different parts of the human body. During optimization, we employ three types of losses \cite{R9, R38}:

\begin{equation}
    \mathcal{L}_{mask} =\|\hat{V}_{mask} - M\|_2
\end{equation}
\begin{equation}
    \mathcal{L}1 =\|\hat{C} - C\| 
\end{equation}
\begin{equation}
    \mathcal{L}_{SSIM} = SSIM(\hat{C}, C)
\end{equation}

where \(\hat{C}\) is the rendered image, \(C\) is the ground truth image, \(\hat{V}_{mask}\) is the cumulative bulk density within the human mask area, \(M\) is a binary number mask label for the ground truth image. Our final loss function is :
\begin{equation}
    \mathcal{L} = \lambda_1\mathcal{L}_{mask} + \lambda_2\mathcal{L}1 + \lambda_3\mathcal{L}_{SSIM}
\end{equation}

\(\lambda\) is the loss weight ratio, while in this experiment we empirically set \(\lambda_1 = 0.3\), \(\lambda_2 = 0.8\) and \(\lambda_3 = 0.1\). During the optimization process, loss-based gradient and opacity Gaussian density adjustments are performed every 400 iterations. The training of our model takes only 1.2K iterations and seamlessly achieves performance comparable to previous methods \cite{R12, R13} in less than 60 seconds on a single GeForce RTX 4090 GPU.

\begin{table*}[h]
\centering
 \resizebox{\linewidth}{!}{
        \begin{tabular*}{\textwidth}{@{\hspace{10pt}}@{\extracolsep{\fill}}c | c c c c c | c c c c c}
        \hline
        Zju$\_$mocap & \multicolumn{5}{c|}{\centering My$\_$377} & \multicolumn{5}{c}{My$\_$386}   \\ \hline
        Metrics & PSNR\(\uparrow\) & SSIM\(\uparrow\) & LPIPS\(\downarrow\) & Train\(\downarrow\) & Pts\(\downarrow\) & PSNR\(\uparrow\) & SSIM\(\uparrow\) & LPIPS\(\downarrow\) & Train\(\downarrow\) & Pts\(\downarrow\) \\ \hline
        HUGS \cite{R13} & 30.7911 &0.97 &0.02 & 2.4m & - & 33.7255 &0.98 &0.03 & 2.5m & - \\ 
        
        GauHuman \cite{R12} & 32.2191 &0.97 &0.03 & 81.3s & 12651 & 33.7288 & 0.96 & 0.03 & 79.9s & 13251 \\  \hline
        Ours & 31.6193 & 0.97 & 0.03 & \textbf{59.8s} & \textbf{10441} & 33.5271 & 0.96 & 0.04 & \textbf{58.5s} & \textbf{8294} \\ \hline
        \(\Delta\) & 0.5998\(\downarrow\) & 0 & 0.01\(\uparrow\) & \textbf{21.5s\(\downarrow\)} & \textbf{2210\(\downarrow\)} & 0.2017\(\downarrow\) & 0.02\(\downarrow\) & 0.01\(\uparrow\) & \textbf{21.4s\(\downarrow\)} & \textbf{4957\(\downarrow\)} \\ \hline
        
        \end{tabular*}
    }
    \\[10pt]
    \resizebox{\linewidth}{!}{
        \begin{tabular*}{\textwidth}{@{\hspace{10pt}}@{\extracolsep{\fill}}c | c c c c c | c c c c c}
        \hline
        Zju$\_$mocap & \multicolumn{5}{c|}{\centering My$\_$387} & \multicolumn{5}{c}{My$\_$392}   \\ \hline
        Metrics & PSNR\(\uparrow\) & SSIM\(\uparrow\) & LPIPS\(\downarrow\) & Train\(\downarrow\) & Pts\(\downarrow\) & PSNR\(\uparrow\) & SSIM\(\uparrow\) & LPIPS\(\downarrow\) & Train\(\downarrow\) & Pts\(\downarrow\) \\ \hline
        HUGS \cite{R13} &29.2681 &0.97 &0.03 & 2.5m & - & 31.3198 &0.97 &0.03 & 2.5m & - \\ 
        
        GauHuman \cite{R12} & 28.2111 &0.95 &0.03 & 81.9s &\hspace{4pt}9896\hspace{4pt} & 32.2488 &0.96 & 0.03 & 81.3s & 11180 \\  \hline
        Ours & 28.2112 & 0.95 & 0.04 & \textbf{61.1s} &\hspace{4pt}\textbf{9808}\hspace{4pt} &31.9332 & 0.96 & 0.04 & \textbf{58.5s} & \textbf{9849} \\ \hline
        \(\Delta\) & 1.0569\(\uparrow\) & 0.02\(\downarrow\) & 0.01\(\uparrow\) & \textbf{20.8s\(\downarrow\)} &\hspace{4pt}\textbf{88\(\downarrow\)} & 0.3156 \(\downarrow\)& 0.01\(\downarrow\) & 0.01\(\uparrow\) & \textbf{22.8s\(\downarrow\)} & \textbf{1331\(\downarrow\)} \\ \hline
        \end{tabular*}
    }
    \\[10pt]
\resizebox{\linewidth}{!}{
    \begin{tabular*}{\textwidth}{@{\hspace{10pt}}@{\extracolsep{\fill}}c | c c c c c | c c c c c}
    \hline
    Zju$\_$mocap & \multicolumn{5}{c|}{\centering My$\_$393} & \multicolumn{5}{c}{My$\_$394}   \\ \hline
    Metrics & PSNR\(\uparrow\) & SSIM\(\uparrow\) & LPIPS\(\downarrow\) & Train\(\downarrow\) & Pts\(\downarrow\) & PSNR\(\uparrow\) & SSIM\(\uparrow\) & LPIPS\(\downarrow\) & Train\(\downarrow\) & Pts\(\downarrow\) \\ \hline
    HUGS \cite{R13} & 29.7898 &0.97 &0.03 & 2.4m & - & 30.5576 &0.97 &0.03 & 2.4m & - \\ 
    
    GauHuman \cite{R12} & 30.2125 &0.96 &0.03 & 82.4s & 10988 & 31.4199 &0.96 & 0.03 & 81.7s & 12783 \\  \hline
    Ours & 30.1190 & 0.95 & 0.04 & \textbf{57.8s} & \textbf{8550} & 31.1025 & 0.95 & 0.04 & \textbf{65.9s} & \textbf{10717} \\ \hline
    \(\Delta\) & 0.0935\(\downarrow\) & 0.02\(\downarrow\) & 0.01\(\uparrow\) & \textbf{24.6s}\(\downarrow\) & \textbf{2438\(\downarrow\)} & 0.3174\(\downarrow\) & 0.02\(\downarrow\) & 0.01\(\uparrow\) & \textbf{15.8s\(\downarrow\)} & \textbf{2066\(\downarrow\)} \\ \hline
    \end{tabular*}
    }

    \caption{Comparison of EfficientHuman $($ours$)$ with HUGS and GauHuman on the Zju$\_$mocap my$\_$377, my$\_$386, my$\_$387, my$\_$392, my$\_$393, my$\_$394 dataset \cite{R37} under 1.2K iterations. Performance is evaluated on PSNR, SSIM, LPIPS, Training Time (Train), and Pts denotes the final number of Gaussians during training. The row of \(\Delta\) represents the absolute difference between our method and the best baseline results (HUGS or GauHuman) for each metric. Notably, our work improves training speed by approximately $30\%$ and reduces redundant Gaussians by around $17\%$ compared to the fastest time-efficient model, GauHuman. Each method is measured on a single RTX 4090.}
    \label{tab:tab1}
\end{table*}

\section{Experiment}
\label{sec:experiment}
\textbf{Comparison Methods.} We compare our method with two state-of-the-art models for dynamic human reconstruction, GauHuman \cite{R12} and HUGS \cite{R13}. For the qualitative results, we select six different themes from the ZJU-MoCap dataset \cite{R37} and demonstrate that EfficientHuman excels in modeling arms and clothing, ensuring the integrity of rendered images. Quantitative results indicate that our method achieves the shortest training time and utilizes the fewest Gaussian points, concurrently sustaining high performance in metrics such as Peak Signal-to-Noise Ratio (PSNR), Structural Similarity Index (SSIM) \cite{R38}, and Learned Perceptual Image Patch Similarity (LPIPS) \cite{R41}. Furthermore, we conduct several ablation studies to substantiate the efficacy of our design modules.

\subsection{Dataset}
\label{ssec:subhead} 
\textbf{ZJU-MoCap Dataset} \cite{R37} is a dataset of a human body photographed by a multi-view camera in the laboratory. We selected six sub-datasets (377, 386, 387, 392, 393, 394) themed on six different actions, with the data captured by 1 camera as training and the remaining 22 cameras as evaluation. The parameters of the cameras, the SMPL\cite{R10} parameters of the human body, and the segmentation are provided directly by the dataset to be able to use it directly.

\subsection{Qualitative Results}
\label{ssec:subhead} 
Although our quantitative evaluation shows a slightly lower Peak Signal-to-Noise Ratio (PSNR) compared to the current state-of-the-art time-efficient model, GauHuman \cite{R12}, this metric alone does not fully capture the visual quality improvements. Our method exhibits two significant qualitative enhancements evident upon visual inspection. Specifically, our rendering results exhibit marked improvements in two critical aspects of 3D human modeling:

1.\hspace{2pt}\textbf{Enhanced Boundary and Detail Preservation}: Our method ensures the integrity and well-defined boundaries on clothing, this improvement is particularly notable in areas such as the hem of shirts and the outlines of coats. Meanwhile, as shown by the red boxes in the sixth to eighth columns of the third row of \figref{fig:fig3}, our approach retains the details of the middle button on the clothing as seen in the real image. The enhanced edge and detail preservation contribute significantly to the overall visual fidelity of the 3D models, creating a more realistic and accurate representation of fabric and clothing dynamics.

2.\hspace{2pt}\textbf{Superior Contour Recovery}: We achieve remarkable advancements in recovering and rendering the contours of human arms. In the red-boxed area of the third row of \figref{fig:fig3}, it is evident that our approach provides superior clarity in the arm contours, as demonstrated in the first two columns and the last two columns. This enhancement captures the subtle curves and edge definition of the arms and the shape of the fist, leading to a more lifelike and anatomically accurate depiction of the human form.

These notable enhancements stem from the direct mapping of Articulated 2D Gaussians onto the dynamic surfaces of the human body within our framework, this fitting approach minimizes the gaps between Gaussians, ensuring a more precise and nuanced depiction of the human form.

\subsection{Quantitative Results}
\label{ssec:subhead} 
\tabref{tab:tab1} shows a comparison between EfficientHuman and 3D Gaussian Splatting (3DGS)-based reconstructions of 3D human models, including GauHuman \cite{R12} and HUGS \cite{R13}, on the ZJU-MoCap dataset \cite{R37}. The evaluation metrics encompass Peak Signal-to-Noise Ratio (PSNR), Structural Similarity Index (SSIM) \cite{R38}, Learned Perceptual Image Patch Similarity (LPIPS) \cite{R41}, Training Time (Train), and the final Gaussian count (Pts) post-training. \(\Delta\) denotes the absolute difference between our method and the best baseline results (HUGS or GauHuman) for each metric.

The quantitative results indicate that our approach achieves high performance in PSNR, SSIM, and LPIPS, while also improving the training speed by approximately $30\%$ and reducing redundant Gaussians by around $17\%$ compared to the fastest time-efficient model \cite{R12}. In another words, our proposed Articulated 2D Gaussian method effectively fits a freely moving human body and converges rapidly, EfficientHuman is on average 20 seconds faster than the fastest time-efficient human reconstruction model (GauHuman), completing training in just 1.2K iterations and approaching state-of-the-art performance in under one minute. The substantial decrease in Gaussian count post-training underscores our method's efficiency in dynamically fitting human body surfaces and eliminating unnecessary Gaussians. To our knowledge, our approach stands out as the most time-efficient for human reconstruction, balancing swift training with the effective pruning of redundant Gaussians while maintaining excellence on critical metrics.

\subsection{Ablation Experiments}
\label{ssec:subhead} 
\tabref{tab:tab2} shows our ablation experiments performed over 1.2K iterations on a subset of 394 in the ZJU-MoCap dataset\cite{R37}. We remove the Linear Blend Skinning (LBS) \cite{R10} optimization and pose calibration components from the complete model and evaluate them using four metrics: Peak Signal-to-Noise Ratio (PSNR), Structural Similarity Index (SSIM) \cite{R38}, Learned Perceptual Image Patch Similarity (LPIPS) \cite{R41} and Training Time (Train). Furthermore, we examine the impact of incorporating the human mask loss \(\mathcal{L}_{mask}\) on performance enhancement with minimal additional time expenditure.

\textbf{LBS Optimization.} The ablation experiment analysis, as shown in \tabref{tab:tab2}, reveals that removing the LBS optimization module results in a $2.04\%$ decrease in PSNR, dropping from 31.1025 to 30.4680. Additionally, SSIM decreases by 0.01, and LPIPS increases from 0.04 to 0.05. Although the training time is reduced by $\approx 10\%$, all other performance metrics show a decline when compared to the metrics of the full model. These variations demonstrate that the LBS optimization module is crucial to our model's performance.

\textbf{Pose Calibration.} When the pose calibration module is omitted, SSIM and LPIPS remain unchanged compared to the full model, while the training time is reduced by $13.63\%$, from 66 seconds to 57 seconds. Notably, PSNR decreases by $0.58\%$, dropping from 31.1025 to 30.9227, indicating that the pose calibration module contributes to the improvement in PSNR.

\textbf{Human Mask Loss \(\mathcal{L}_{mask}\).} Excluding \(\mathcal{L}_{mask}\) results in $\approx 2\%$ reduction in PSNR, decreasing from 31.1025 to 30.4826, this reflects the degradation of image detail and quality. While SSIM remains stable at 0.95 and the training time is reduced by approximately $7.58\%$, LPIPS increases by $25\%$, from 0.04 to 0.05, which suggests a lower perceived image quality. \(\mathcal{L}_{mask}\) is crucial for preserving the visual quality and detail in image reconstruction and should not be overlooked.

\textbf{Full Model.} The complete model delivers superior performance across all metrics, with a slight increase in training time. Its SSIM value of 0.95 underscores the model's effectiveness in preserving structural integrity during reconstruction, crucial for tasks reliant on spatial relationships and image patterns. The model's minimal LPIPS score of 0.04 further confirms its high quality, as LPIPS correlates well with human perception, indicating that the reconstructed images closely resemble the originals. This excellence comes with only a marginal training time increase, demonstrating the efficiency of our approach. The incremental $\approx 6$ seconds is a minor trade-off for the significant enhancement in reconstruction quality, essential for practical applications demanding both speed and accuracy.

Each ablation study underscores the individual contributions of the components, including LBS optimization and pose calibration. Furthermore, \(\mathcal{L}_{mask}\) plays a critical role in maintaining or enhancing model performance, albeit with a minor impact on training time
 \begin{table}[]
    \centering
    \begin{tabularx}{\linewidth}{l c c c c}
    \hline
        Zju$\_$mocap & \multicolumn{4}{c}{My$\_$394} \\ \hline
        Metrics & PSNR\(\uparrow\) & SSIM\(\uparrow\) & LPIPS\(\downarrow\) & \hspace{-5pt}Train\(\downarrow\) \\ \hline
        
        {\footnotesize w/o LBS optimization} & 30.4680 & 0.94 & 0.05 & \hspace{-5pt}59s \\ \hline
        {\footnotesize w/o pose calibration} & 30.9227 & \textbf{0.95} & \textbf{0.04} & \hspace{-5pt}\textbf{57s} \\ \hline
        {\footnotesize w/o \(\mathcal{L}_{mask}\)} & 30.4826 & \textbf{0.95} & 0.05 & \hspace{-5pt}61s \\ \hline
        {\footnotesize Ours(full model)} & \textbf{31.1025} & \textbf{0.95} & \textbf{0.04} & \hspace{-5pt}66s \\ \hline
    \end{tabularx}
    \caption{ Quantitative results of ablating LBS optimization, pose calibration and \(\mathcal{L}_{mask}\)  are evaluated on PSNR, SSIM, LPIPS and Training Time (Train).}
    \label{tab:tab2}
\end{table}

\section{Conclusion}
\label{sec:conclusion}
In summary, our work aims to expedite training and uphold rendering quality in 3D human modeling through the use of Articulated 2D Gaussians. Our primary objectives are twofold: to substantially decrease the computational training time without compromising rendering quality and to eliminate redundant Gaussian elements from the final model. Our experiments on the ZJU-MoCap dataset demonstrate that the model achieves training in under a minute with only 1.2K iterations, reducing the Gaussian count by approximately $17\%$ compared to the fastest existing time-efficient model \cite{R12}. This reduction not only accelerates the training process but also refines the details of clothing and human arm contours. Our work exemplifies a significant advancement in 3D human modeling, providing a robust and time-efficient alternative to existing techniques.\\
\textbf{Limitation and Future Work.} While EfficientHuman delivers rapid training and commendable rendering quality, it faces constraints imposed by the human prior model SMPL \cite{R10}, particularly in capturing fine details of clothing textures and finger restoration. Our work centers on the efficient reconstruction of 3D humans using Articulated 2D Gaussians, with the framework also facilitating the extraction of mesh from 2D splats. Thus, Signed Distance Fields (SDF) \cite{R14, R42} for mesh extraction present a promising avenue for future work. This lays a foundation for high-resolution texture mapping, enabling the precise depiction of clothing textures. Further improvements to the final model's performance can be achieved through post-processing, including texture enhancement, smoothing, and the incorporation of detailed clothing features.

\vfill\pagebreak

{\small
\bibliographystyle{ieee_fullname}
\bibliography{egbib}

\begin{thebibliography}{10}\itemsep=-1pt

\bibitem{R30}
Bharat~Lal Bhatnagar, Cristian Sminchisescu, Christian Theobalt, and Gerard Pons-Moll.
\newblock Loopreg: Self-supervised learning of implicit surface correspondences, pose and shape for 3d human mesh registration.
\newblock {\em Advances in Neural Information Processing Systems}, 33:12909--12922, 2020.

\bibitem{R24}
Jianchuan Chen, Ying Zhang, Di Kang, Xuefei Zhe, Linchao Bao, Xu Jia, and Huchuan Lu.
\newblock Animatable neural radiance fields from monocular rgb videos.
\newblock {\em arXiv preprint arXiv:2106.13629}, 2021.

\bibitem{R31}
Xu Chen, Yufeng Zheng, Michael~J Black, Otmar Hilliges, and Andreas Geiger.
\newblock Snarf: Differentiable forward skinning for animating non-rigid neural implicit shapes.
\newblock In {\em Proceedings of the IEEE/CVF International Conference on Computer Vision}, pages 11594--11604, 2021.

\bibitem{R55}
Yue Chen, Xuan Wang, Xingyu Chen, Qi Zhang, Xiaoyu Li, Yu Guo, Jue Wang, and Fei Wang.
\newblock Uv volumes for real-time rendering of editable free-view human performance.
\newblock In {\em Proceedings of the IEEE/CVF Conference on Computer Vision and Pattern Recognition}, pages 16621--16631, 2023.

\bibitem{R1}
Hongsuk Choi, Gyeongsik Moon, Matthieu Armando, Vincent Leroy, Kyoung~Mu Lee, and Grégory Rogez.
\newblock Mononhr: Monocular neural human renderer.
\newblock In {\em 2022 International Conference on 3D Vision (3DV)}, pages 242--251, 2022.

\bibitem{R17}
Alvaro Collet, Ming Chuang, Pat Sweeney, Don Gillett, Dennis Evseev, David Calabrese, Hugues Hoppe, Adam Kirk, and Steve Sullivan.
\newblock High-quality streamable free-viewpoint video.
\newblock {\em ACM Transactions on Graphics (ToG)}, 34(4):1--13, 2015.

\bibitem{R35}
Boyang Deng, John~P Lewis, Timothy Jeruzalski, Gerard Pons-Moll, Geoffrey Hinton, Mohammad Norouzi, and Andrea Tagliasacchi.
\newblock Nasa neural articulated shape approximation.
\newblock In {\em Computer Vision--ECCV 2020: 16th European Conference, Glasgow, UK, August 23--28, 2020, Proceedings, Part VII 16}, pages 612--628. Springer, 2020.

\bibitem{R18}
Mingsong Dou, Sameh Khamis, Yury Degtyarev, Philip Davidson, Sean~Ryan Fanello, Adarsh Kowdle, Sergio~Orts Escolano, Christoph Rhemann, David Kim, Jonathan Taylor, et~al.
\newblock Fusion4d: Real-time performance capture of challenging scenes.
\newblock {\em ACM Transactions on Graphics (ToG)}, 35(4):1--13, 2016.

\bibitem{R58}
Jiemin Fang, Taoran Yi, Xinggang Wang, Lingxi Xie, Xiaopeng Zhang, Wenyu Liu, Matthias Nie{\ss}ner, and Qi Tian.
\newblock Fast dynamic radiance fields with time-aware neural voxels.
\newblock In {\em SIGGRAPH Asia 2022 Conference Papers}, pages 1--9, 2022.

\bibitem{R48}
Sara Fridovich-Keil, Alex Yu, Matthew Tancik, Qinhong Chen, Benjamin Recht, and Angjoo Kanazawa.
\newblock Plenoxels: Radiance fields without neural networks.
\newblock In {\em Proceedings of the IEEE/CVF conference on computer vision and pattern recognition}, pages 5501--5510, 2022.

\bibitem{R22}
Xiangjun Gao, Jiaolong Yang, Jongyoo Kim, Sida Peng, Zicheng Liu, and Xin Tong.
\newblock Mps-nerf: Generalizable 3d human rendering from multiview images.
\newblock {\em IEEE Transactions on Pattern Analysis and Machine Intelligence}, 2022.

\bibitem{R43}
Stephan~J Garbin, Marek Kowalski, Matthew Johnson, Jamie Shotton, and Julien Valentin.
\newblock Fastnerf: High-fidelity neural rendering at 200fps.
\newblock In {\em Proceedings of the IEEE/CVF international conference on computer vision}, pages 14346--14355, 2021.

\bibitem{R6}
Chen Geng, Sida Peng, Zhen Xu, Hujun Bao, and Xiaowei Zhou.
\newblock Learning neural volumetric representations of dynamic humans in minutes.
\newblock In {\em Proceedings of the IEEE/CVF Conference on Computer Vision and Pattern Recognition}, pages 8759--8770, 2023.

\bibitem{R16}
Kaiwen Guo, Peter Lincoln, Philip Davidson, Jay Busch, Xueming Yu, Matt Whalen, Geoff Harvey, Sergio Orts-Escolano, Rohit Pandey, Jason Dourgarian, et~al.
\newblock The relightables: Volumetric performance capture of humans with realistic relighting.
\newblock {\em ACM Transactions on Graphics (ToG)}, 38(6):1--19, 2019.

\bibitem{R44}
Peter Hedman, Pratul~P Srinivasan, Ben Mildenhall, Jonathan~T Barron, and Paul Debevec.
\newblock Baking neural radiance fields for real-time view synthesis.
\newblock In {\em Proceedings of the IEEE/CVF international conference on computer vision}, pages 5875--5884, 2021.

\bibitem{R3}
Shoukang Hu, Fangzhou Hong, Liang Pan, Haiyi Mei, Lei Yang, and Ziwei Liu.
\newblock Sherf: Generalizable human nerf from a single image.
\newblock In {\em Proceedings of the IEEE/CVF International Conference on Computer Vision (ICCV)}, pages 9352--9364, October 2023.

\bibitem{R12}
Shoukang Hu, Tao Hu, and Ziwei Liu.
\newblock Gauhuman: Articulated gaussian splatting from monocular human videos.
\newblock In {\em Proceedings of the IEEE/CVF Conference on Computer Vision and Pattern Recognition}, pages 20418--20431, 2024.

\bibitem{R14}
Binbin Huang, Zehao Yu, Anpei Chen, Andreas Geiger, and Shenghua Gao.
\newblock 2d gaussian splatting for geometrically accurate radiance fields.
\newblock In {\em ACM SIGGRAPH 2024 Conference Papers}, pages 1--11, 2024.

\bibitem{R7}
Tianjian Jiang, Xu Chen, Jie Song, and Otmar Hilliges.
\newblock Instantavatar: Learning avatars from monocular video in 60 seconds.
\newblock In {\em Proceedings of the IEEE/CVF Conference on Computer Vision and Pattern Recognition}, pages 16922--16932, 2023.

\bibitem{R9}
Bernhard Kerbl, Georgios Kopanas, Thomas Leimk{\"u}hler, and George Drettakis.
\newblock 3d gaussian splatting for real-time radiance field rendering.
\newblock {\em ACM Trans. Graph.}, 42(4):139--1, 2023.

\bibitem{R13}
Muhammed Kocabas, Jen-Hao~Rick Chang, James Gabriel, Oncel Tuzel, and Anurag Ranjan.
\newblock Hugs: Human gaussian splats.
\newblock In {\em Proceedings of the IEEE/CVF conference on computer vision and pattern recognition}, pages 505--515, 2024.

\bibitem{R54}
Georgios Kopanas, Julien Philip, Thomas Leimk{\"u}hler, and George Drettakis.
\newblock Point-based neural rendering with per-view optimization.
\newblock In {\em Computer Graphics Forum}, volume~40, pages 29--43. Wiley Online Library, 2021.

\bibitem{R20}
Youngjoong Kwon, Dahun Kim, Duygu Ceylan, and Henry Fuchs.
\newblock Neural human performer: Learning generalizable radiance fields for human performance rendering.
\newblock {\em Advances in Neural Information Processing Systems}, 34:24741--24752, 2021.

\bibitem{R56}
Youngjoong Kwon, Lingjie Liu, Henry Fuchs, Marc Habermann, and Christian Theobalt.
\newblock Deliffas: Deformable light fields for fast avatar synthesis.
\newblock {\em Advances in Neural Information Processing Systems}, 36, 2024.

\bibitem{R53}
Christoph Lassner and Michael Zollhofer.
\newblock Pulsar: Efficient sphere-based neural rendering.
\newblock In {\em Proceedings of the IEEE/CVF Conference on Computer Vision and Pattern Recognition}, pages 1440--1449, 2021.

\bibitem{R45}
Ruilong Li, Hang Gao, Matthew Tancik, and Angjoo Kanazawa.
\newblock Nerfacc: Efficient sampling accelerates nerfs.
\newblock In {\em Proceedings of the IEEE/CVF International Conference on Computer Vision}, pages 18537--18546, 2023.

\bibitem{R28}
Haotong Lin, Sida Peng, Zhen Xu, Yunzhi Yan, Qing Shuai, Hujun Bao, and Xiaowei Zhou.
\newblock Efficient neural radiance fields for interactive free-viewpoint video.
\newblock In {\em SIGGRAPH Asia 2022 Conference Papers}, pages 1--9, 2022.

\bibitem{R10}
Matthew Loper, Naureen Mahmood, Javier Romero, Gerard Pons-Moll, and Michael~J Black.
\newblock Smpl: A skinned multi-person linear model.
\newblock {\em Seminal Graphics Papers: Pushing the Boundaries, Volume 2}, pages 851--866, 2023.

\bibitem{R34}
Marko Mihajlovic, Yan Zhang, Michael~J Black, and Siyu Tang.
\newblock Leap: Learning articulated occupancy of people.
\newblock In {\em Proceedings of the IEEE/CVF Conference on Computer Vision and Pattern Recognition}, pages 10461--10471, 2021.

\bibitem{R5}
Ben Mildenhall, Pratul~P Srinivasan, Matthew Tancik, Jonathan~T Barron, Ravi Ramamoorthi, and Ren Ng.
\newblock Nerf: Representing scenes as neural radiance fields for view synthesis.
\newblock {\em Communications of the ACM}, 65(1):99--106, 2021.

\bibitem{R8}
Thomas M{\"u}ller, Alex Evans, Christoph Schied, and Alexander Keller.
\newblock Instant neural graphics primitives with a multiresolution hash encoding.
\newblock {\em ACM transactions on graphics (TOG)}, 41(4):1--15, 2022.

\bibitem{R42}
Richard~A Newcombe, Shahram Izadi, Otmar Hilliges, David Molyneaux, David Kim, Andrew~J Davison, Pushmeet Kohi, Jamie Shotton, Steve Hodges, and Andrew Fitzgibbon.
\newblock Kinectfusion: Real-time dense surface mapping and tracking.
\newblock In {\em 2011 10th IEEE international symposium on mixed and augmented reality}, pages 127--136. Ieee, 2011.

\bibitem{R29}
Sida Peng, Shangzhan Zhang, Zhen Xu, Chen Geng, Boyi Jiang, Hujun Bao, and Xiaowei Zhou.
\newblock Animatable neural implicit surfaces for creating avatars from videos.
\newblock {\em arXiv preprint arXiv:2203.08133}, 4(5), 2022.

\bibitem{R37}
Sida Peng, Yuanqing Zhang, Yinghao Xu, Qianqian Wang, Qing Shuai, Hujun Bao, and Xiaowei Zhou.
\newblock Neural body: Implicit neural representations with structured latent codes for novel view synthesis of dynamic humans.
\newblock In {\em Proceedings of the IEEE/CVF Conference on Computer Vision and Pattern Recognition}, pages 9054--9063, 2021.

\bibitem{R52}
Ruslan Rakhimov, Andrei-Timotei Ardelean, Victor Lempitsky, and Evgeny Burnaev.
\newblock Npbg++: Accelerating neural point-based graphics.
\newblock In {\em Proceedings of the IEEE/CVF Conference on Computer Vision and Pattern Recognition}, pages 15969--15979, 2022.

\bibitem{R47}
Christian Reiser, Songyou Peng, Yiyi Liao, and Andreas Geiger.
\newblock Kilonerf: Speeding up neural radiance fields with thousands of tiny mlps.
\newblock In {\em Proceedings of the IEEE/CVF international conference on computer vision}, pages 14335--14345, 2021.

\bibitem{R51}
Darius R{\"u}ckert, Linus Franke, and Marc Stamminger.
\newblock Adop: Approximate differentiable one-pixel point rendering.
\newblock {\em ACM Transactions on Graphics (ToG)}, 41(4):1--14, 2022.

\bibitem{R21}
Shunsuke Saito, Zeng Huang, Ryota Natsume, Shigeo Morishima, Angjoo Kanazawa, and Hao Li.
\newblock Pifu: Pixel-aligned implicit function for high-resolution clothed human digitization.
\newblock In {\em Proceedings of the IEEE/CVF international conference on computer vision}, pages 2304--2314, 2019.

\bibitem{R33}
Shunsuke Saito, Jinlong Yang, Qianli Ma, and Michael~J Black.
\newblock Scanimate: Weakly supervised learning of skinned clothed avatar networks.
\newblock In {\em Proceedings of the IEEE/CVF Conference on Computer Vision and Pattern Recognition}, pages 2886--2897, 2021.

\bibitem{R36}
Johannes~L Schonberger and Jan-Michael Frahm.
\newblock Structure-from-motion revisited.
\newblock In {\em Proceedings of the IEEE conference on computer vision and pattern recognition}, pages 4104--4113, 2016.

\bibitem{R15}
Johannes~L Sch{\"o}nberger, Enliang Zheng, Jan-Michael Frahm, and Marc Pollefeys.
\newblock Pixelwise view selection for unstructured multi-view stereo.
\newblock In {\em Computer Vision--ECCV 2016: 14th European Conference, Amsterdam, The Netherlands, October 11-14, 2016, Proceedings, Part III 14}, pages 501--518. Springer, 2016.

\bibitem{R39}
Noah Snavely, Steven~M Seitz, and Richard Szeliski.
\newblock Photo tourism: exploring photo collections in 3d.
\newblock {\em ACM siggraph 2006 papers}, pages 835--846, 2006.

\bibitem{R26}
Shih-Yang Su, Frank Yu, Michael Zollh{\"o}fer, and Helge Rhodin.
\newblock A-nerf: Articulated neural radiance fields for learning human shape, appearance, and pose.
\newblock {\em Advances in neural information processing systems}, 34:12278--12291, 2021.

\bibitem{R19}
Zhuo Su, Lan Xu, Zerong Zheng, Tao Yu, Yebin Liu, and Lu Fang.
\newblock Robustfusion: Human volumetric capture with data-driven visual cues using a rgbd camera.
\newblock In {\em Computer Vision--ECCV 2020: 16th European Conference, Glasgow, UK, August 23--28, 2020, Proceedings, Part IV 16}, pages 246--264. Springer, 2020.

\bibitem{R38}
Zhou Wang, Alan~C Bovik, Hamid~R Sheikh, and Eero~P Simoncelli.
\newblock Image quality assessment: from error visibility to structural similarity.
\newblock {\em IEEE transactions on image processing}, 13(4):600--612, 2004.

\bibitem{R25}
Hongyi Xu, Thiemo Alldieck, and Cristian Sminchisescu.
\newblock H-nerf: Neural radiance fields for rendering and temporal reconstruction of humans in motion.
\newblock {\em Advances in Neural Information Processing Systems}, 34:14955--14966, 2021.

\bibitem{R32}
Ze Yang, Shenlong Wang, Sivabalan Manivasagam, Zeng Huang, Wei-Chiu Ma, Xinchen Yan, Ersin Yumer, and Raquel Urtasun.
\newblock S3: Neural shape, skeleton, and skinning fields for 3d human modeling.
\newblock In {\em Proceedings of the IEEE/CVF conference on computer vision and pattern recognition}, pages 13284--13293, 2021.

\bibitem{R49}
Alex Yu, Ruilong Li, Matthew Tancik, Hao Li, Ren Ng, and Angjoo Kanazawa.
\newblock Plenoctrees for real-time rendering of neural radiance fields.
\newblock In {\em Proceedings of the IEEE/CVF International Conference on Computer Vision}, pages 5752--5761, 2021.

\bibitem{R50}
Qiang Zhang, Seung-Hwan Baek, Szymon Rusinkiewicz, and Felix Heide.
\newblock Differentiable point-based radiance fields for efficient view synthesis.
\newblock In {\em SIGGRAPH Asia 2022 Conference Papers}, SA '22, New York, NY, USA, 2022. Association for Computing Machinery.

\bibitem{R41}
Richard Zhang, Phillip Isola, Alexei~A Efros, Eli Shechtman, and Oliver Wang.
\newblock The unreasonable effectiveness of deep features as a perceptual metric.
\newblock In {\em Proceedings of the IEEE conference on computer vision and pattern recognition}, pages 586--595, 2018.

\bibitem{R4}
Fuqiang Zhao, Wei Yang, Jiakai Zhang, Pei Lin, Yingliang Zhang, Jingyi Yu, and Lan Xu.
\newblock Humannerf: Generalizable neural human radiance field from sparse inputs.
\newblock {\em arXiv preprint arXiv:2112.02789}, 3:1, 2021.

\bibitem{R57}
Zerong Zheng, Xiaochen Zhao, Hongwen Zhang, Boning Liu, and Yebin Liu.
\newblock Avatarrex: Real-time expressive full-body avatars.
\newblock {\em ACM Transactions on Graphics (TOG)}, 42(4):1--19, 2023.

\end{thebibliography}
}

\end{document}